# Road Detection Technique Using Filters with Application to Autonomous Driving System

Yinka O. Agunbiade, Johnson O. Dehinbo, Tranos Zuva and Adeyinka K Akanbi

*Abstract*— Autonomous driving systems are broadly used equipment in the industries and in our daily lives; they assist in production, but are majorly used for exploration in dangerous or unfamiliar locations. Thus, for a successful exploration, navigation plays a significant role. Road detection is an essential factor that assists autonomous robots achieved perfect navigation. Various techniques using camera sensors have been proposed by numerous scholars with inspiring results, but their techniques are still vulnerable to these environmental noises: rain, snow, light intensity and shadow. In addressing these problems, this paper proposed to enhance the road detection system with filtering algorithm to overcome these limitations. Normalized Differences Index (NDI) and morphological operation are the filtering algorithms used to address the effect of shadow and guidance and re-guidance image filtering algorithms are used to address the effect of rain and/or snow, while dark channel image and specular-to-diffuse are the filters used to address light intensity effects. The experimental performance of the road detection system with filtering algorithms was tested qualitatively and quantitatively using the following evaluation schemes: False Negative Rate (FNR) and False Positive Rate (FPR). Comparison results of the road detection system with and without filtering algorithm shows the filtering algorithm's capability to suppress the effect of environmental noises because better road/non-road classification is achieved by the road detection system with filtering algorithm. This achievement has further improved path planning/region classification for autonomous driving system.

## I. INTRODUCTION

Injuries and accidents are common occurrences in our environment. In vehicle for instance, numerous situations can cause injuries and accidents, but behavioural factors and human decision are the major contributions to these fatality rates [1]. Figure 1 presents the percentage contribution of various behavioural factors that can cause accidents with overtaking contributing the least and over speeding contributing the most to this fatality. In 2002, it was estimated that 20-50 million people are injured globally from vehicle accident and an average of 1.2 million people died from the same fatality [2].

Manuscript received September 9, 2016; revised September 28, 2016.
Yinka Olusanya Agunbiade is with the Department of Computer Science, Tshwane University of Technology, Soshanguve, South of Pretoria, South Africa.
Johnson Olumuyiwa Dehinbo is with the Department of Computer Science, Tshwane University of Technology, Soshanguve, South of Pretoria, South Africa
Tranos Zuva is with the Department of Computer Science, Vaal University of Technology, Vander Bay, Free State, South Africa
Adeyinka Kabir Akanbi is with the Department of Information Technology, Central University of Technology, South Africa

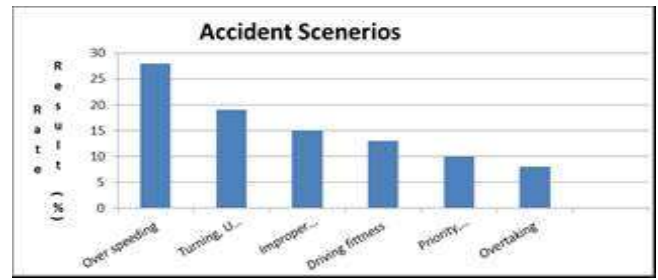

Fig I: Behavioural factors contribution to accident [1].

In South Africa from the report of 2001-2006, it was estimated that the road transportation system contributes the most (99.8%), to the total transport accident deaths [2]. However, to reduce injuries and accidents, autonomous driving system could be used to assist drivers in making better decisions or to drive a vehicle without being controlled by anybody. The Autonomous driving system needs an effective vision system to detect correctly road region for perfect navigation [3] and sensors are used to achieve vision. Sensors are categorized into active and passive ones, they are used to extract information from the environment that will be interpreted by the road detection system of the autonomous driving system [4]. Some scientists [5, 6] propose the use of laser scanner sensor based on elevation variance for road detection. The elevation variance for extracting plain surfaces which represent road region will be low, while the elevation variance for extracting boundary which separate the road from non-road will be high, but only limited to unstructured environments and can gives wrong measurement on black objects that are not reflecting light [7] or shinny surfaces/objects. Other scientists like [8, 9] propose the use of radar sensors for road detection. The radar sensor method is based on 5 state vectors illustrated in equation 1.

$$\hat{x} = \begin{pmatrix} {}^SC & {}^VO & {}^S\gamma & {}^SW & {}^SS \end{pmatrix}^T \quad (I)$$

where ${}^SC$ implies the curvature, ${}^VO$ symbolises distance of the vehicle to the road centre, ${}^S\gamma$ indicates the orientation angle of the road, ${}^SW$ signifies the road width, ${}^SS$ characterises lane offset, the radar distance between left most lane mark and the left road boundary.

Equation 1 is used to design the road ahead of the autonomous driving system. The road geometry result is remarkable, but the sensor is short-sighted and expensive [10]. In some situations, certain scientists [11, 12] propose the use of sonic sensors for road extraction and detection. The range information of sonic sensors for regions that are empty represent road while those that are occupied represent non-road. Thus, the range information of sonic sensors is inconsistent because in some situations, empty areas might not necessarily be a road region, but can perform better on unstructured environment [6]. Recently, camera is used as a sensor for providing vision for autonomous driving systems to overcome some of the limitations encounter by scientists using active sensors to develop their technique. However, environmental noises are factors that lead to poor navigation performance of the autonomous driving system using camera sensors and these support why some researchers still prefer the use of active sensors for vision [9]. In an attempt to propose a better road detection system for autonomous driving system, using camera sensor, study revealed that environmental noises have the capability to affect the color properties of the image with significant effects of miss-classification of road as non-road and vice versa [13]. In this research, filters are introduced to minimize the effect of environmental noises for proper classification of road/non-road region to be attained.

## II. Literature Review

Road region detection is a significant study, attracting the attention of researchers because of its support towards navigation [14]. Navigation on the other hand facilitates autonomous robot movement and can cause robot accidents if proper road is not detected for navigation. In achieving proper navigation, several researchers has proposed many techniques, this section reviews some of these researchers, their techniques and the limitations they encountered during the implementation of their systems.

In [15], they proposed the use of camera sensor to improve the road detection of autonomous driving system in various ways, since their awareness of environmental noises. They planned to address light intensity by proposing the use of $HSV$ color space, because in $HSV$ representation light intensity effect is reduced. This implies that there will be less misclassification in the presence of light intensity, to further improve the road detection; texture technique which measure the local spatial variation in image intensity was combined with color as signified as $F_{i\,j}$ in equation 2.

$$F_{i\,j} = \left[ f_{t_1(i,j)},..,f_{t_5(i,j)}, f_{c_1(i,j)},..,f_{c_3(i,j)} \right] \quad (2)$$

where,
$i = 1,.......,H \quad j = 1,.......,W$

$f_{t_n(i,j)}$ represents $n^{th}$ haralick statistical feature at point $(i, j)$, $f_{c_n(i,j)}$ denotes $n^{th}$ color feature at point $(i, j)$ in HSV space, $H$ and $W$ represent respectively height and width of the image.

In this technique, five statistical haralick and three colors features are combined to form eight element feature vector for road detection. The next operation in their technique is the use of the segmentation algorithm. At this stage, Support Vector Machine (SVM) is employed. SVM classifier performs two important operations; it is used for training data and also for data classification. Morphological operation was introduced for connecting component region and hole filling while online learning operation is used for a comparison between classification result and morphological operation. The comparison operation evaluates the quality of the SVM classifier and generates new training set from the morphological result. This operation improves the adaptability of the system and reduces the possibility of misclassification of road as non-road and verse versa, even in the presence of light intensity. However, the system for road detection only targets light intensity, but can fail when it encounters other environmental noises like rain, snow and shadow. In [16], visual navigation is considered to be important for autonomous robot and they propose the use of camera sensor. Their technique is color and edge based. At first they propose the conversion of Red $(R)$, Green $(G)$ and Blue $(B)$ space to $HSV$, before conversion, pixel might have similar value of $RGB$ with an effect on the Hue value of such pixel to be uncertain. In addressing this issue, they introduced a concept of color purity $(CP)$ as expressed in equation 3.

$$CP = \frac{S \cdot V}{255} = \frac{(I_{max} - I_{min})}{225} \quad (II)$$

where $S$ is denoted as saturation, $V$ is represented as value, $I_{max} = \max(R,G,B)$ and $I_{min} = \min(R,G,B)$.

Since the approach involves color, a selected area needs to be processed and classified to be road or non-road region. Judging whether pixel of selected region belongs to an object or not, various distance classification measures exist, which range from Euclidean distance, minkowski distance and chebyster distance, but in the research, manhanthan distance from point $A$ to point $B$ represented as $d(i, road)$ in equation 4 and thresholding algorithm is used for pixel classification.

$$d(i, road) = a.|I_H - E_H| + b.|I_{CP} - E_{CP}| \quad (4)$$

where $I_H$ denotes all hue component pixels in the selected area, $E_H$ represents the mean of all selected pixels' hue component. $I_{CP}$ signifies color purity in the selected road region and $E_{CP}$ represents the mean of all selected pixels' color purity.

The next operation is edge point detection. In solving this problem, Least Square Method (LSM) represented as $K$ in equation 5 was proposed for edge detection and fitting.

$$K = \sqrt{\sum_{i=0}^{n} \left[ f(x_i) - p(x_i) \right]^2} \rightarrow \min \quad (5)$$

where $p(x)$ signifies fitting function, $f(x)$ represents original function and $(x_i, f(x_i)), (i = 1,2,,,,,n)$ represents the data point available.

In their experiment, remarkable result was achieved because they combined both color and edge technique. This helps the system to cope with shadow and light intensity, but with slow processing speed because it is computationally intensive, and fails if water area is too large and during rain fall.

[17] Proposed the use of camera sensor to address the issue of road detection using classification algorithm. In their technique, K-means algorithm was introduced to detect cluster center for background and rain pixel. The distance between cluster center and pixel intensity represented as $d(I_p, w_b)$ is given in equation 6

$$d(I_p, w_b) = |I_p - w| \qquad (6)$$

where $I_p$ represents the pixel intensity value and $w$ signifies cluster center. In K-means technique, image pixel is classified as background using equation 7 otherwise it is rain pixel.

$$d(I_p, w_b) < d(I_p, w_r) \qquad (7)$$

where $w_b$ and $w_r$ denote background cluster and rain pixel.

After the detection technique process, color blending approach is applied to candidate rain pixel classified by equation 7. This technique replaces rain pixel color with corresponding background color and this generates a new color pixel $(C)$. Illustration is expressed in equation 8.

$$C = \alpha C_b + (1-\alpha)C_r \qquad (8)$$

where $C_r$ signifies rain affected color and $C_b$ represents background color.

After the removal operation is the application of the classification algorithm which is used to judge if a pixel belongs to the class of road or non-road. This technique can only preform impressively in rain scenarios. However, with other environmental noises the system will fail to detect road region correctly.

However, according to the work of [18] it was reviewed that there is problem associated to road/lane detection under different weather conditions (snow, rain, fog and mist). The effect makes image interpretation impossible with miss-classification effect [14] because different regions will be having similar characteristics. Thus, since our technique also involve colours, our aim of achieving better road detection even in the presence of these noises. Therefore, filtering algorithms will be introduced to minimise the effect of these environmental noises (shadow, rain, light intensity and snow) for perfect Navigation to be attained.

## III. METHODOLOGY

In an attempt to develop an efficient and effective system, this section explains in full, the component and assumptions that are employed to enhance the road detection system to achieve better classification result of road/non-road. The road detection system consists of four stages: Feature extraction stage, Filtering Stage, Segmentation Stage and Morphological Stage. Figure 2 illustrate schematically the flow of operation of these stages.

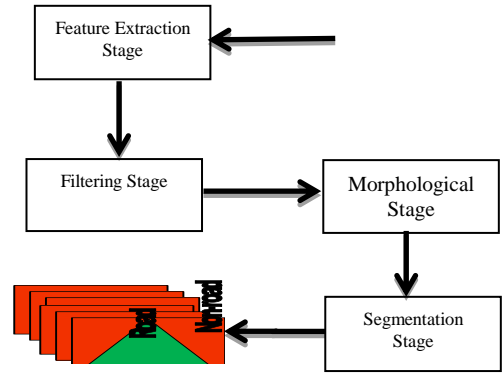

Fig 2: Road detection system for autonomous driving system

### A. Feature Extraction

The feature extraction stage represents the first stage of the road detection system. In this phase, image feature for various regions are extracted (road region, non-road region and uncertainty) using various filters and statistical technique are extracted from the captured images [19]. There are various types of feature that are valuable in image processing: color, texture and edge. These features are vital components for proposing a remarkable recognition system. In this study, color is the selected feature and it has been a widely used technique for road detection [20, 21]. Color is directly available as $RGB$ intensity from camera and possess a unique way for object recognition, which can be employed for road detection based on some certain assumptions and rules [15]. The assumption that road has similar color dissimilar from non-road has been proposed by [22] with outstanding results. However, using $RGB$ color technique has a limitation of poor classification when facing environmental noises. Environmental noises have abilities to interrupt the assumption with miss-classification effect because various regions will be having comparable color characteristics.

### B. Filtering Stage

Environmental noises have the ability to degrade image, bring poor vision and corrupt the $RGB$ color component valve which results to poor extraction of road and non-road feature extraction. Filtering algorithm stage is the second phase of the road region detection system and was introduced to the system to suppress the effect of environmental noise affecting the feature extraction stage for better classification result to be achieved at segmentation phase. Three filtering algorithms were introduced to detect and remove the environmental noises such as rain, snow, light intensity and shadow only. Illustrations of the functionality for these filtering algorithms on images were discussed in this section. In the Shadow algorithm, $RGB$ color code is converted to $HSV$ because shadow holds easy identification with maximum value of saturation $(S)$ and the minimum Value $(V)$, in $HSV$ color space. Illustrations for the conversion of $RGB$ to $HSV$ is given in Equation 10 – 12 [16].

$$V = \frac{1}{3}(R + G + B) \qquad (10)$$

$$S = 1 - \frac{3}{(R + G + B)} \min(R, G, B) \qquad (11)$$

$$H = \begin{cases} \theta & \text{if } B \leq G \\ 360^\circ & \text{if } B > G \end{cases} \qquad (12)$$

where

$$\theta = \cos^{-1}\left\{ \frac{\frac{1}{2}[(R-G)+(R-B)]}{\sqrt{(R-G)^2 + (R-B)(G-B)}} \right\},$$

$RGB$ is characterized as red, green, blue. The value $(V)$ and saturation $(S)$ components of I are used to extract shadow area. Equation 13 illustrates the use of Normalized Differences Index $(NDI)$.

$$NDI = \frac{S - V}{S + V} \qquad (13)$$

NDI images are segmented by the use of OTSU thresholding algorithm to find an optimal threshold $(T)$ [23]. Pixels with higher NDI value than threshold $(T)$ are categorized as shadow pixel else non-shadow. After thresholding, $I_{shadow}(i, j)$ indicated as binary image with pixel of shadow set to 1, and pixel of non-shadow is set to 0. In the removal process of shadow, connected component algorithm is employed to connect region categorised as 1. In the removal process of shadow, the transformation function is represented as $I_k^i(i, j)$ has expressed in equation 18. This is the mean and variance of the buffer area used to compensate the shadow regions.

$$I_k^i(i, j) = \mu_{buff,k} + \frac{I_k(i, j) - \mu_K}{\sigma_{buff,k}} \qquad (18)$$

where $\mu_{buff,K}$ and $\sigma_{buff,K}$ represents the mean and variance of the pixels of image I at location $I_{buff,K}$, while $\mu_K$ and $\sigma_K$ represents the mean and variance of the shadow pixels of image I at location $I_K$ [23].

### I. RAIN AND FILTERING ALGORITHM

This algorithm is capable of minimizing the effect of snow and rain. The snow and rain removal method is based on modelling first intensity value pixel of image captured on a rain or snow day when passing through an element on the charge coupled devices (CCDs) of the camera. $I_{rs}$ as expressed in equation 19 symbolizes intensity value of the image captured during rain or snow.

$$I_{rs} = \alpha I_E + (1 - \alpha) I_b \qquad (19)$$

thus, $I_E$ symbolizes the intensity valve of a stationary drop of snowflakes keep still at time when captured by the camera, $I_b$ symbolizes the background intensity. Let $\alpha = \frac{\tau}{T}$ is the variable used where $T$ is the time of exposure and the time taken $\tau$ raindrop or snowflake is passing through an element on the CCD.

The removal process of rain and snow is based on the equation in but, unfortunately some important information such as edges that has similar color with either raindrop or snowflakes are removed alongside with rain and snow.

$$J_i = \bar{a}_k I_i + \bar{b}_k \qquad (21)$$

where

$$a_k = \left(\left(\sum_{i \in \omega_k} I_i P_i\right) / |\omega| - \mu_k \bar{p}_k\right) / (\sigma_k^2 + \varepsilon),$$

$$b_k = \bar{P}_k - a_k \mu_k$$

i symbolizes the pixel location in I, $\omega_k$ is the kernel region, k symbolises location in the kernel, $|\omega|$ represents a number of pixels in $\omega_k$, $\sigma_k^2$ and $\mu_k^2$ represents the variance and mean of the input image I respectively in $\omega_k$ and $P_k$ represents mean value of P in $\omega_k$.

However, the important information removed together with rain and snow by guided filter is restored by using equation 22 to extract a redefined guidance image $(I_g)$ with no rain and snow effect.

$$I_g = (I_f + J_g)/2 \qquad (22)$$

where $J_g$ symbolizes the gray-image of $J_i - I_{rs}$ [24].

### II. LIGHT FILTERING ALGORITHM

The light filtering algorithm has the ability to minimize the effect of light intensity caused by sun light and this is often common in our society because of high intensity of light generated from the sun. The removal method is carried out by modelling object reflected by color camera based on dichromatic reflection model which represent the linear combination of diffuse reflection components and specular represented as $I(x)$ is expressed in equation 23.

$$I(x) = I^D(x) + I^s(x) = w_d(x) B(x) + w_s(x) G \qquad (23)$$

where, $I$ symbolizes the observed image intensity, $x = \{x, y\}$ symbolizes the image coordinate, $I^D$ symbolizes diffuse reflection component, $I^S$ symbolizes specular reflection component, $B(x)$ symbolizes the diffuse color, G symbolizes specular color, $w_d(x)$ symbolizes the coefficient that governs the magnitude of diffuse reflection component and $w_s(x)$ symbolizes the coefficient that governs the magnitude of specular reflection component.

The light intensity detection method proposed is based on the use of dark channel $(I^{dark}(x))$ shown in equation 24 and automatic thresholding is used to find high light reflection in image. The combination of these methods is employed for appropriate classification of regions of image affected with light intensity. The concept behind light intensity detection is based on regions with affected high light, will have high intensity value while regions that are not affected will have low intensity value in the dark channel model [25].

$$I^{dark}(x) = \min_{y \in v(x)} \left( \min_{c \in (r,g,b)} (I^c(y)) \right) \quad (24)$$

where $v(x)$ denotes the local patch centered at $x$, $x$ represents the image coordinate and $I^c$ signifies color channel of $I$.

The marked image produced by automatic thresholding of dark channel image is labelled as 1, which signifies the region affected by light intensity and 0 to represent regions that are not affected.

The method for minimizing the effects of light intensity is based on specular-to-diffuse proposed by [25]. Illustration of image with minimal light intensity effect $(I^D)$ is expressed in equation 27

$$I^D(\wedge_{max}) = I - \frac{\max_{u \in (r,g,b)} I_u - \wedge_{max} \sum_{u \in (r,g,b)} I_u}{1 - 3\wedge_{max}} \quad (27)$$

where $\wedge_{max}$ represents the specular pixels, $r, g$ and $b$ represent the corresponding histogram components of red, green and blue.

Thus, after the procedures of these filters, the environmental noise influences on images is minimized and corrupted color properties of affected regions in images are improved. This enhancement makes the system understand the image content properly that will facilitate classification at the next stage.

*C. Image Segmentation Stage*

Image segmentation is an important procedure in computer vision. This represents the third stage of the road detection system. Image segmentation is employed to sub-divide digital image into numerous regions [14]. More specifically, they are used to label pixels in images for a visual characteristic based on pixel having similar labels. Hence, this characteristic can be employed to extract objects and boundaries in images, this segmentation stage was used to assist the system to achieve better extraction and classification of road /non-road region since the corrupted pixel properties of the image caused by environmental noises are minimised. However, in the segmentation stage, Support Vector Machine (SVM) was employed as our segmentation algorithm because of improved performance in generalization ability [26]. In SVM, lagrange multiplier for linearly separable of an image the optimization problem is illustrated in 29.

$$\text{Maximize}_{\lambda_1 \ldots \lambda_l} \sum_{i=1}^{l} \lambda_i - \frac{1}{2} \sum_{i=1}^{l} \sum_{i=1}^{l} \lambda_i \lambda_j y_i y_j x_i . x_j \quad (29)$$

Subjected to constraint

$$\sum_{i=1}^{l} \lambda_i y_i = 0, \lambda_i \geq 0 \quad , \quad i = 1\ldots l$$

The use of decision function illustrated in equation 30 is applied for pattern recognition and classification.

$$F(x) = \text{sign}\left( \sum_{i=1}^{l} y_i \lambda_i (x.x_i) + b \right) \quad (30)$$

where $i$ signifies images pixels, $\lambda_1 \ldots \lambda_l$ is symbolized as the vector of non-negative Lagrange multiplier corresponding the constraint in equation 29. The vector $x_i, y_i$ when $\lambda_i > 0$ are termed support vector, others are training vectors when $\lambda_i = 0$ [27].

*D. Post Processing Stage*

The Post processing procedure is the fourth stage of the road detection system. This is a vital phase and morphological operation is employed to carry out this operation because it is used for geometrical description of image content based on set rules, matrix rules, surface structure and random function [17, 28-30]. It's a common algorithm used by earlier researchers. In morphological operation, it is assumed that road areas are connected and related together. Based on this assumption, connected component algorithms illustrated in equation 31 was employed for the extraction of largest connected road area, eroding away also holes in the connected road region. Other regions dissimilar from the connected road regions are categorized as non-road.

$$X_k = (X_{k-1} \oplus B) \cap A^c, k = 1, 2, 3 \ldots \quad (31)$$

where $X_0 = P$ (initial point), $B$ is symbolized as symmetric structure element, $\cap$ signifies interception operator, $A^c$ means the complement of set image $A$, $X_K$ contain all the filled holes. At iteration $K$ the algorithm ends if $X_k = X_{k-1}$. The set union of $A$ holds the edge [29].

IV. EXPERIMENTAL EVALUATION AND RESULT

In this paper, the enhanced road detection system with three filters was tested using several images affected with snow, rain, shadow and light intensity. These real life images were obtained by the use of camera and some were obtained via internet because they are unavailable for the camera to capture. The experimental performance is carried out by qualitative and quantitative method. Illustrations thereof are given in experiments 4.1, and 4.2.

*Experiment 4.1: Qualitative results of the road detection system.*

The experiment employed here is to observe the performance of the enhanced road detection system qualitatively. This qualitative experiment conducted was between the road detection system with and without filters using the same real life frame of images. This will help to reveal the road detection system with best performance. Illustrations of the result attained for these experiments are given below. Figure 3a, 4a, 5a and 6a signifies real life images used for the experiment, Figure 3b, 4b, 5b and 6b signifies the result of the road detection system without filtering algorithm whereas, Figure 3c, 4c, 5c and 6c signifies the result of the road detection system with filters

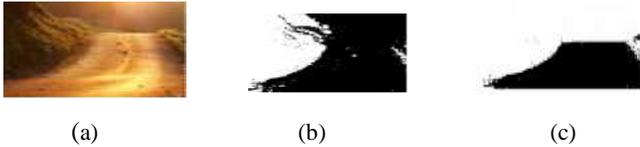

(a)      (b)      (c)

Fig 3: Road image (a) signify the real life image with light intensity noise, image (b) and (c) respectively represent the vision results for the road detection system without and with filters.

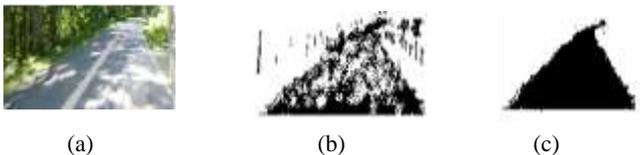

(a)      (b)      (c)

Fig 4: Road image (a) signify the real life image mostly suffering from shadow and a bit lighting variation noises, Image (b) and (c) respectively represent vision results for the road detection system without and with filters.

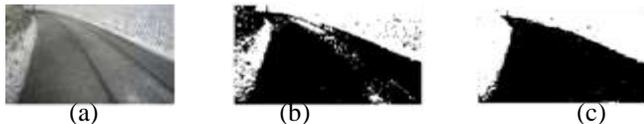

(a)      (b)      (c)

Fig 5: Road image (a) signify the real life image with snow noise, image (b) and (c) respectively represent the vision results for the road detection system without and with filters

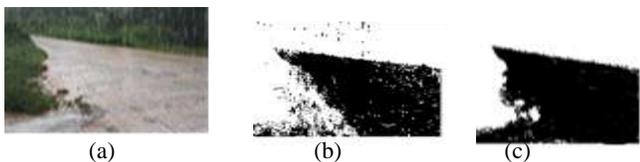

(a)      (b)      (c)

Fig 6: Road image (a) signify the real life image suffering from rain noises, Image (b) and (c) respectively represent are vision results for the road detection system without and with filters.

The experiment for this section reveals that figure 3b, 4b, 5b and 6b which represent the result of the road detection system without filters has many road region misclassified as non-road and vice versa, but Figure 3c, 4c, 5c and 6c which represent the result of the road detection system with filters as less road region misclassified as non-road and vice versa. These experiments support the significant of filters in road detection systems.

*Experiment 4.2: Quantitative result of the road detection system.*

In this section, the experiment employed was carried out quantitatively and comparison was still between the road detection system with and without filters. In the literature, various quantitative evaluation schemes exist, but the ones used in this study were expressed below in equation 32–33. These evaluation schemes are employed to judge pixels of qualitative result based on ground truth classification. Thus, FNR signifies the false negative rate and FPR symbolizes false positive rate.

$$\text{FNR} = \frac{FN_i}{TP_i + FN_i}, \qquad (32)$$

The ratio of non-roads pixels that are classified properly,

$$\text{FPR} = \frac{FP_i}{TN_i + FP_i}, \qquad (33)$$

The ratio of non-road pixels that are classified as road, where $FP_i$ means false positive: numbers of non-road pixels classified as road in the $i^{th}$ image. $FN_i$ symbolizes false negative: numbers of road pixel that are labelled as non-road in the $i^{th}$ image and $N_i$ means numbers of road and non-road pixels in $i^{th}$ ground truth classification.

Using these experimental schemes or evaluation measures, the qualitative result for 30 frames of images were tested at an average of 3 frames for accuracy and error. The results attained are used for a comparison between the system with and without filtering algorithms. The graphical representation result in Figure 7a shows the illustration of the experiment and comparison result reveals that the road region detection system with filtering algorithms achieves better performance than the system without filters because minimum error rate is achieved in all cases. This clearly supports that the filters introduced to enhance the road detection system are capable of minimising the effect of environmental noises.

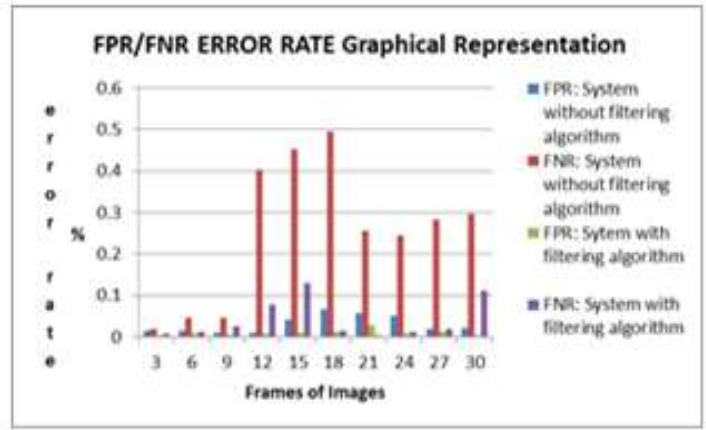

Fig 7a: FPR/FNR Error Rate Result

## V. CONCLUSION

This study proposed a method to improve road detection system for autonomous driving system in an outdoor environment affected with environmental noises (snow, rain, light intensity and shadow). In our road detection technique, the filters employed for improving the performance of the road detection system assist in minimizing the effect of these environmental noises. The post processing stage (morphological operation) proposed in our road detection system assists in accomplishing a better classification result. Experimental performance is carried out by employing real life images suffering from various snow, rain, light variation intensity and shadow noises. The results obtained support the system enhancement in road detection and this accomplishment has further improved autonomous navigation for robots. Although, the road detection system was successfully enhanced with filters that minimizes the effect of environmental noises, unfortunately the computational complexity of the system is high with an effect that will slow down the motion of the autonomous driving system because of decrease in image processing speed. In future, this problem will be addressed by introducing parallel algorithm.